\documentclass[11pt]{article}
\usepackage{amsmath,amssymb,amscd,amsthm,etoolbox}
\usepackage{graphicx}
\usepackage[pdftex,colorlinks]{hyperref}
\usepackage[margin=1.0in]{geometry}
\usepackage{enumerate}
\usepackage{mathtools}

\usepackage{newtxmath}

\usepackage{algorithm}
\usepackage{algorithmic}
\usepackage{xcolor, tcolorbox}

\usepackage[round]{natbib}
\hypersetup{citecolor=blue}

\usepackage[normalem]{ulem}

\makeatletter
\let\original@footnote\footnote
\newcommand{\align@footnote}[1]{%
  \ifmeasuring@
    \chardef\@tempfn=\value{footnote}%
    \footnotemark
    \setcounter{footnote}{\@tempfn}%
  \else
    \iffirstchoice@
      \original@footnote{#1}%
    \fi
  \fi}
\pretocmd{\start@align}{\let\footnote\align@footnote}{}{}
\makeatother

\newtheorem{theorem}{Theorem}[section]
\newtheorem{lemma}[theorem]{Lemma}
\newtheorem{remark}[theorem]{Remark}

\newtheorem{counter-example}[theorem]{Counter example}
\newtheorem{proposition}[theorem]{Proposition}

\newtheorem{open question}[theorem]{Open question}
\newtheorem{corollary}[theorem]{Corollary}

\newtheorem{definition}[theorem]{Definition}

\newtheorem{claim}{Claim}
\newtheorem{prop}{Proposition}
\newtheorem*{claim*}{Claim}

\newcommand{\eps}{\epsilon}

\DeclareMathOperator{\Err}{L_D}
\DeclareMathOperator{\Errs}{L_D^*}
\DeclareMathOperator{\EmpS}{\hat{L}_S}

\DeclareMathOperator{\ErrS}{\hat{L}_S^*}
\newcommand{\R}{\mathtt{LR}}
\newcommand{\G}{\mathtt{LG}}
\newcommand{\cost}{c}
\newcommand{\unif}{e}

\DeclareMathOperator*{\E}{\mathbb{E}}

\newcommand{\new}[1]{\textcolor{black}{#1}}

\title{Monotone Learning}

\author{
Olivier Bousquet
\and
Amit Daniely
\and
Haim Kaplan
\and
Yishay Mansour
\and
Shay Moran
\and
Uri Stemmer
}

\begin{document}
\maketitle

\thispagestyle{empty}

\begin{abstract}
The amount of training-data is one of the key factors which determines 
    the generalization capacity of learning algorithms. 
    Intuitively, one expects the error rate to decrease 
    as the amount of training-data increases.
    Perhaps surprisingly, natural attempts to formalize this intuition give rise to 
    interesting and challenging mathematical questions. 
    For example, in their classical book on pattern recognition, 
    \cite*{devroye1996probabilistic} ask whether there exists a {monotone} Bayes-consistent algorithm.
    This question remained open for over 25 years, 
    until recently \cite*{pestov2021universally} 
    resolved it for binary classification,
    using an intricate construction of a monotone Bayes-consistent algorithm.

We derive a general result in multiclass classification, showing that 
    \emph{every} learning algorithm~$A$ can be transformed to a monotone one with similar performance. 
    Further, the transformation is efficient and only uses a black-box oracle access to~$A$. 
    This demonstrates that one can provably avoid non-monotonic
    behaviour without compromising performance, thus answering
    questions asked by \cite*{devroye1996probabilistic},
    \cite*{viering2019open}, \cite*{Viering21curves}, and by \cite*{mhammedi2021risk}.
    
Our general transformation readily implies monotone learners in a variety of contexts:
    for example, Pestov's result follows by applying it on \emph{any} 
    Bayes-consistent algorithm (e.g., $k$-Nearest-Neighbours).
    In fact, our transformation extends Pestov's result 
    to classification tasks with an \emph{arbitrary} number of labels. 
    This is in contrast with Pestov's work which is tailored to binary classification. 

In addition, we provide \emph{uniform bounds} on the error of the monotone algorithm.
    This makes our transformation applicable in distribution-free settings.
    For example, in PAC learning it implies that \emph{every}
    learnable class admits a monotone PAC learner.
    This resolves questions asked by~\cite*{viering2019open,Viering21curves,mhammedi2021risk}.
\end{abstract}

\section{Introduction}
In this work we study the following fundamental question.
Pick some standard learning algorithm~$A$, and consider training it 
    for some natural task using a data-set of $n$ examples.
\begin{center}
{Does feeding $A$ with more training data \emph{provably} reduces its population loss?}    
\end{center}
    E.g., would increasing the number of examples from $n$ to $n+1$ improve its loss?
    How about~$2n$? Can one guarantee improvement in this case?
    What about~$2^n$, or even $2^{2^n}$? Would that be sufficient?
    Can one at least assure that the loss will not deteriorate?

Intuitively, the answer should be yes:
    indeed, the more often we face a certain task, 
    the better we typically get at solving it.
    This basic intuition is reflected in many works in theoretical 
    and applied machine learning.
    For example, \cite*{SSbook} assert in their book
    that the learning curve starts decreasing when the number of examples surpasses the VC dimension (page~153); \cite*{DudaHartStork01} state in their book that for real-world problems the learning curve is monotone (Subsection 9.6.7). 
    Similar statements are made by a variety of other works, a partial list includes~\cite*{Gu01perf,Tax08curves,Weiss2014GeneratingWL}.
    We refer the reader to the thorough survey by \cite*{Viering21curves}  for an extensive discussion about monotone and non-monotone learning curves (Section 6).

On the other hand, one might argue that in order to successfully learn
    complex functions, the algorithm must dedicate time and resources to exploring
    larger and larger sets of hypotheses, and consequently exhibit a non-monotone
    behaviour. For example, any Bayes consistent learning algorithm
    must consider arbitrarily complex hypotheses 
    (since it is able to approximate arbitrary functions).
    Indeed, one can show that consistent algorithms such as \emph{Nearest-Neighbors}
    can demonstrate such non-monotone behaviour \citep*{devroye1996probabilistic}.
    This intuition is reflected in Chapter~6 in the book by \cite*{devroye1996probabilistic} in which it is conjectured 
    that no Bayes-consistent rules can be monotone  (Problem~6.16).

\vspace{2mm}

These intuitive considerations inspire a host of theoretical questions.
Is it really the case that ``more data $=$ better generalization''? 
Is it at least the case for natural algorithms and natural learning tasks? 
Perhaps it is too much to expect that the addition of a single example
will lead to better performance, but maybe if one doubles the training-set 
then better performance is guaranteed? Can we at least guarantee
that the performance does not deteriorate?

\paragraph{Notation.}
We focus on multiclass classification with respect to the zero/one loss
    and use standard learning theoretic notation (see, e.g., \cite*{SSbook}).
    Let $X$ be a set called the domain and let $Y$ denote the label-set.
    We assume that $Y$ is finite, w.l.o.g $Y=[k]=\{0,1,\ldots, k-1\}$.
    For a set $Z$, let $Z^\star :=\cup_{n=0}^\infty Z^n$ denote the space of all finite sequences with elements from $Z$. An {\em hypothesis (or classifier)} is a function $h:X\to Y$. 
    An {\em example} is a pair $z=(x,y)\in X\times Y$.
    A {\em sample} $S\in (X\times Y)^\star$ is a (finite) sequence of examples.
    A {\em learning rule} is a mapping from $(X\times Y)^\star$ to $Y^X$;
    i.e., the input is a finite sample and the output is a hypothesis.
    Given a distribution $D$ over $X\times Y$ and a hypothesis $h$, 
    the {\em (population) loss} of $h$ with respect to $D$ 
    is $\Err(h) = \E_{(x,y)\sim D}1[h(x)\neq y]$.
    Given a sample $S=\{(x_i,y_i)\}_{i=1}^m$, 
    the {\rm (empirical) loss} of~$h$ with respect to $S$
    is  $\EmpS(h) =\frac{1}{m}\sum_{i=1}^m{1[h(x_i)\neq y_i]}$, where $1[\cdot]$ is the indicator function.

\subsubsection*{Monotone Learners}
\begin{definition}[Monotone learning rule]\label{def:mon}
A learning rule $M$ is said to be \emph{monotone} w.r.t a distribution~$D$
if,
\[(\forall m):\E_{S\sim D^{m}}\Bigl[\Err\bigl(M(S)\bigr)\Bigr] \geq \E_{S\sim D^{m+1}}\Bigl[\Err\bigl(M(S)\bigr)\Bigr].\]
That is, the expected population loss of $M$ is monotone non-decreasing
in the size of its training set.
\end{definition}

The following theorem is the main result in this work; it asserts that every learning algorithm  can be efficiently transformed to a monotone one with competitive generalization guarantees.

\begin{theorem}\label{t:main}[Every learner can be monotonised]
Consider the setting of multiclass classification to $k\in\mathbb{N}$ labels.
Then, every learning algorithm $A$ can be efficiently converted to a learning algorithm $M=M(A)$ such that $M$ has  only a black-box oracle access to~$A$ and
\begin{enumerate}
    \item $M$ is monotone with respect to \emph{every} distribution $D$.
    \item $M$'s performance is \emph{competitive} with that of $A$:
        for every source distribution $D$,
    \[\Bigl(\forall m\Bigr)\Bigl(\exists m' \text{ s.t. } \frac{m}{30}-1\leq m'\leq m\Bigr):\E_{S\sim D^{m}}\Bigl[\Err\bigl(M(S)\bigr)\Bigr] \leq \E_{S\sim D^{m'}}\Bigl[\Err\bigl(A(S)\bigr)\Bigr] + O\Bigl(\sqrt{\frac{\log m}{m}}\Bigr).\]
\end{enumerate}
\end{theorem}

Theorem~\ref{t:main} affirmatively answers questions posed 
    by \cite*{devroye1996probabilistic},
    \cite*{viering2019open}, \cite*{Viering21curves}, and \cite*{mhammedi2021risk}.
    In addition, Theorem~\ref{t:main} readily implies monotone learners in a variety of contexts:
    
\begin{corollary}[Bayes-Consistent Monotone Learners]
For every $k\in\mathbb{N}$ there exists a Bayes consistent monotone learner for multiclass classification into $k$ labels.
\end{corollary}
Indeed, this follows by applying the transformation on {\it any} Bayes-consistent learner (for example, $k$-nearest neighbor).
This extends Pestov's result who focused on the case of binary classification
and designed a clever histogram-based Bayes consistent algorithm.
Moreover, while Pestov's algorithm and analysis are tailored to the binary case,
our argument is more general, and at the same time 
conceptually (and arguably technically) simpler.
The existence of Bayes-optimal consistent learner remained open for 25 years
since it was asked by by \cite*{devroye1996probabilistic}.

Theorem~\ref{t:main} is also applicable in other contexts. 
In fact, the distribution-free regret-bound on the learning rate of
the monotone learner allows one to apply it in the PAC setting,
where monotone learners were not known to exist~\cite*{viering2019open,Viering21curves}:
\begin{corollary}[Monotone PAC Learners]
Let $k\in\mathbb{N}$ and let $\mathcal{H}\subseteq [k]^X$ be a PAC learnable hypothesis class. Then, there exists a monotone agnostic PAC learner for $\mathcal{H}$.
\end{corollary}
Indeed, this follows by applying the transformation 
    on any PAC learning algorithm for $\mathcal{H}$ (say any empirical risk minimizer). 
    Note that the learning rate of the resulting monotone PAC learner 
    is suboptimal by an additive $\log m$ factor\footnote{The optimal PAC learning rate is proportional to $\sqrt{1/m}$.}.
    We leave the exploration for the optimal monotone PAC learning rate to future work.

\subsection{Informal Explanation}
A learning rule, even if it is Bayes consistent, does not have any reason, a priori, to be monotone. Indeed, the fact that the expected error converges to the Bayes error does not mean the convergence happens monotonically and it could very well be that the error strictly increases between $m$ and~$m+1$ infinitely often.
Let us examine what are the difficulties one would encounter when trying to convert a Bayes consistent learner into one that is monotone.

\paragraph{Convergent Learners are Sparsely Monotone.}
Given any learning algorithm $A$ that has the following convergence property:
\[
\lim_{m\to\infty} \E_{s\sim D^m}\left[\Err(A(S))\right] = e_D,
\]
with $\E_{s\sim D^m}\left[\Err(A(S))\right]\ge e_D$ for infinitely many $m$, which for example happens for any Bayes-consistent learner, we immediately get that there exists a subsequence $m_{1},\ldots,m_{k},\ldots$ of indices over which the error will be monotonically decreasing, i.e.,
\[
\forall j\ge i,\,
\E_{s\sim D^{m_{j}}}\left[\Err(A(S))\right] \le \E_{s\sim D^{m_{i}}}\left[\Err(A(S))\right].
\]
However, the above subsequence of indices is \emph{distribution-dependent}, which means that we cannot guarantee that if we increase the sample size from some value $m$ to some other value $m'$, the error will be smaller for all distributions.

\paragraph{Sparse to Dense Monotonicity.}
We first observe that we could relax the monotonicity requirement to hold only for infinitely many steps (or arbitrary size). Indeed, while the monotonicity requirement of Definition \ref{def:mon} is written for $m$ and~$m+1$, we observe that if we had a learner that satisfies a \emph{sparse} version of this inequality, such as
\begin{equation}\label{eq:sparse-mon}
    (\forall m)(\exists m'\ge m):\E_{S\sim D^{m}}\Bigl[\Err\bigl(M(S)\bigr)\Bigr] \geq \E_{S\sim D^{m'}}\Bigl[\Err\bigl(M(S)\bigr)\Bigr]\,,
\end{equation}
it would be easy to convert it into a learner that satisfies Definition \ref{def:mon} without affecting the limit of its population loss as $m\to\infty$.  Indeed, the above condition guarantees that there is an infinite sequence $m_1,\ldots,m_k,\ldots$ of indices over which the algorithm is guaranteed not to increase its expected loss. This sequence can be defined as $m_1=1$ and $m_{k+1}=m'(m_k)$. And from this, one could create a learner that, given $m$ examples with $m\in [m_k,m_{k+1})$, simply ignores $m-m_k$ examples from the training set. The monotonicity condition would then be satisfied with equality between $m_k$ and~$m_{k+1}$ since the output of our algorithm would be unchanged (in expectation).

\paragraph{Running over Prefixes.}
So in order to convert an arbitrary learning rule into one that is monotone, while still retaining its convergence properties, the main idea is to run the algorithm on prefixes of the training sample and measure the loss of the produced hypotheses in order to pick the best one. As the sample size increases, the pool of hypotheses to choose from will increase and the best one from a larger pool will thus have a smaller loss than from a smaller pool.
This idea has been previously explored by~\cite*{viering2020making,mhammedi2021risk}.

However, implementing this idea turns out to be a subtle task.
Indeed, we can only \emph{estimate} the loss of the hypotheses (by setting aside some examples and computing their empirical loss), so there is always some possibility that we choose a worse hypothesis (which empirically looks better).

To illustrate the issue, let's consider the simplest possible situation where the base algorithm has produced a hypothesis $h_0$ on a prefix of the sample, and another hypothesis $h_1$ on a longer prefix. If $\Err(h_1)\le \Err(h_0)$, the base algorithm is already monotone, but in the case $\Err(h_1)> \Err(h_0)$, any wrapper algorithm would have to choose between outputting $h_0$ or $h_1$.
Unfortunately, this choice will necessarily worsen the error (unless the wrapper always outputs $h_0$ deterministically, in which case it would not manage to track the performance of the base algorithm).
Indeed, the expected loss of any wrapper would be a convex combination of $\Err(h_1)$ and $\Err(h_0)$ and would be strictly larger than $\Err(h_0)$.

So we cannot simply take the output of the base algorithm, and the idea is to \emph{regularize} it, i.e., make it possibly a little worse but in such a way that this regularization can be reduced as the sample size increases and thus we can guarantee monotonicity.

\paragraph{How to Make the Learner Worse?}
The question thus becomes: given a hypothesis $h$, is there a way to produce a hypothesis $h'$ that is guaranteed to be \emph{worse} than $h$ (i.e., $\Err(h')>\Err(h)$) and to possibly control how much worse it is? 
The first idea that comes to mind is to sometimes output a label which we know is incorrect. This would be possible if we add some extra label at our disposal, e.g., in binary classification we would allow the learner to output something different from $0$ or $1$, say $\bot$ and count $\bot$ as a mistake. But this is somewhat artificial and would require to change the nature of the algorithm's predictions. 
So the second idea that comes to mind is to just add noise to the output of the algorithm, i.e., to randomly pick a different label than the one predicted. 
Unfortunately, in the context of classification, adding noise does not guarantee that the loss is made worse! Indeed, in binary classification, if $\Err(h)>1/2$, adding some uniform noise to the output would make the error closer to $1/2$, hence better and not worse.

\paragraph{How to Make the Learner Always Better Than Some Value?}
So we see that if we could, given a hypothesis $h$, which could have error larger than $1/2$, return one that is guaranteed to have error less than $1/2$, we could then make the latter worse by adding uniform noise. This brings us to our last key idea: we symmetrize the output by replacing the hypothesis produced by the base algorithm by the best (in terms of empirical error) between $h$ and $1-h$. We can then guarantee (we will prove it below) that the expected loss of this \emph{symmetrized} output is less than $1/2$ and adding noise will thus strictly increase its loss, making room for reducing the loss when we choose between $h_0$ and~$h_1$. 

Symmetrization is more subtle in the context of multiclass classification with $k>2$ labels; the idea there is to replace $h$ with the best out of $k$ hypotheses which are obtained by composing $h$ with a cyclic permutation of the labels. 
For simplicity, we focus on the binary-case in this outline.

\paragraph{Formalization.} Let us now try and write down some of the ideas above more formally.
The overall approach is to run the base algorithm on a prefix of size $n$ of the sample $S$ to obtain some hypothesis~$h_0$, apply some transformation (which we call \emph{regularization}) to $h_0$ which consists of symmetrizing and adding noise, in order to obtain $R(h_0,S)$. Then perform the same operation on a longer prefix  of size $N>n$ of $S$ to obtain $R(h_1,S)$ and then decide whether to use $h_0$ or $h_1$ by estimating their respective errors (on an additional subset of  $N$ examples from $S$).
If we denote by $p_N$ the probability of choosing $h_1$ over $h_0$, we see that the expected error of the resulting procedure will have the form\footnote{We will later provide more details about how to split the training sample so as to guarantee independence and decouple the expecations appropriately.}
\[
p_N  \E_{S\sim D^{N}}\Bigl[\Err\bigl(R(h_1,S)\bigr)\Bigr] + (1-p_N) \E_{S\sim D^{N}}\Bigl[\Err\bigl(R(h_0,S)\bigr)\Bigr] 
\]
and if we want to satisfy Inequality \eqref{eq:sparse-mon}, this quantity would have to be smaller than the expected error of our procedure ran on $n$ examples, i.e., we would want
\[
p_N  \E_{S\sim D^{N}}\Bigl[\Err\bigl(R(h_1,S)\bigr)\Bigr] + (1-p_N) \E_{S\sim D^{N}}\Bigl[\Err\bigl(R(h_0,S)\bigr)\Bigr] \le \E_{S\sim D^{n}}\Bigl[\Err\bigl(R(h_0,S)\bigr)\Bigr]
\]
As discussed above, without regularization, i.e., if $R$ is the identity, there is no way to guarantee this inequality for every pair $h_0,h_1$, and the problematic case is when $h_1$ is worse than $h_0$, or when $R(h_1,S)$ is worse than $R(h_0,S)$.
If we rewrite the above condition as follows:
\[
p_N \left( \E_{S\sim D^{N}}\Bigl[\Err\bigl(R(h_1,S)\bigr)\Bigr] - \E_{S\sim D^{N}}\Bigl[\Err\bigl(R(h_0,S)\bigr)\Bigr] \right) \le \E_{S\sim D^{n}}\Bigl[\Err\bigl(R(h_0,S)\bigr)\Bigr] - \E_{S\sim D^{N}}\Bigl[\Err\bigl(R(h_0,S)\bigr)\Bigr]
\]
we see that in order for it to be satisfied even when $R(h_1,S)$ is worse than $R(h_0,S)$, we need
\begin{enumerate}
    \item $p_N$ has to be small enough (to make the left hand side small enough).
    \item The regularization over $N$ examples has to be strictly better than the regularization over $n$ examples (to make the right hand side positive and large enough).
\end{enumerate}

\subsection{Technical Contributions}
The technical contribution in this work can roughly be partitioned to two parts:
\begin{itemize}
    \item[(i)] We develop a general axiomatic framework for constructing transformations which compile arbitrary learners  to monotone learners with similar guarantees (Section~\ref{sec:general}).
    We attempt to state this framework in an abstract manner with the hope
    that it might be useful for other loss functions.

In a nutshell, this framework reduces the task to constructing for every hypothesis $h$
    a \emph{small} and \emph{symmetric} class $B_h$ such that $h\in B_h$,
    and $B_h$ can be learned by a monotone learner;
    for example, in the context of binary classification ($Y=\{0,1\}$)
    we use $B_h=\{h, 1-h\}$. More generally, in the context of multiclass classification ($Y=[k]=\{0,\ldots,k-1\}$), we use $B_h=\{s_i\circ h : i\in[k]\}$, where $s_i$ is the cyclic permutations mapping a label $y$ to $y+i \mod k$.

\item[(ii)] In Sections~\ref{sec:binary} and \ref{sec:multiclass} we use our general framework 
    to prove Theorem~\ref{t:main}. In Section~\ref{sec:binary} we focus on the
    case of binary classification; this section serves as a warmup
    to the general multiclass setting which is considered in Section~\ref{sec:multiclass}.

The most technical proof in this work is that of Proposition~\ref{prop:multiclass},
specifically Lemma~\ref{le:monmulti} which asserts that the randomized ERM over $B_h$ is monotone:     Recall that for a hypothesis $h:X\to \{0,\ldots, k-1\}$, the class $B_h$
    consists of the $k$ cyclic permutations of $h$: $B_h = \{s_i\circ h : i\in[k]\}$,
    where $s_i$ is a cyclic permutation mapping $y\mapsto y+i\mod k$.
    The randomized ERM is the algorithm which given an input sample $S$,
    outputs an empirical risk minimizer from $B_h$ which is drawn uniformly at random.

To prove Proposition~\ref{prop:multiclass} we exploit the following symmetry exhibited by $B_h$:
    for any example~$(x,y)$ there exists a unique $h'\in B_h$ such that $h'(x)=y$.
    This implies, via a symmetrization argument and via Chebychev's sum inequality\footnote{Chebyshev's sum inequality asserts that if $a_1\leq a_2\leq\ldots  \leq a_n$ and $b_1\geq b_2\geq\ldots b_n$ then $\frac{1}{n}\sum a_i\cdot \frac{1}{n}\sum b_i \geq \frac{1}{n}a_ib_i$.} the desired monotonicity \citep*{hardy1988inequalities}.
    
We also note that the upper bound on the rate in Theorem~\ref{t:main}
    is \emph{independent} of the number of labels $k$.    
    To achieve this we once again appeal to the symmetric structure of $B_h$,
    and show that $B_h$ satisfies uniform convergence with rate which is independent of $k$.
\end{itemize}

\subsection{Related Work}
The idea of monotone learning curves for universally consistent learners was first discussed by \cite*{devroye1996probabilistic}.This problem attracted little attention until recently when ~\cite*{viering2019open} considered monotone learning
    in a variety of contexts (e.g., when the goal is to learn a fix hypothesis class) and under more general loss functions.
    \cite*{viering2020making}, \cite*{Viering21curves}, and \cite*{mhammedi2021risk}
    considered the problem of transforming a given learner to a monotone one using a \emph{wrapper algorithm}.
    \cite*{viering2020making} and \cite*{mhammedi2021risk} derive weaker forms of monotonicity and
    leave open the question of whether such a transformation exists.
    In this work we resolve this problem in the context of multiclass classification.

The conjecture by~\cite{devroye1996probabilistic} was finally answered in the positive by \cite*{pestov2021universally}. Pestov's result applies to binary classification, and here we prove
an extension to general multiclass classification.

\paragraph{Other Notions of Mononoticity.}
\cite*{viering2020making} proposes to relax the requirement of monotonicity in expectation into \emph{high-probability} and \emph{eventual monotonicity}. {\cite*{viering2020making} and \cite*{mhammedi2021risk} also discuss the relationship with the multiple descent phenomenon established for many learners in recent years.}

\paragraph{Other Notions of Consistency.}
It is important to note that the open problem proposed by \cite{viering2019open} is concerning consistency with respect to a fixed class of function. So this is less general than the universal consistency of \citep*{devroye1996probabilistic}.

\paragraph{Comparing with Pestov's result.}
While \cite*{pestov2021universally} just builds a specific algorithm and not a generic wrapper, and considers only the binary classification case, there are some similarities between his approach and ours that are worth illustrating. Indeed, his algorithm consists in the following three ingredients
\begin{enumerate}
    \item Consider prefixes of the input sample of (exponentially) increasing size
    \item Perform a majority vote over a partition of the input domain
    \item Decide (empirically) whether or not to split each element of the partition into smaller pieces
\end{enumerate}
The first ingredient is similar to our (and other's) approach of guaranteeing monotonicity on an infinite sequence of indices (what we call sparse monotonicity above), the second one bears some similarity with our symmetrization approach since the majority vote consists in comparing $h$ and $1-h$, and the last one is comparable with our update procedure which decides whether to continue using $h_0$ or to switch to $h_1$.

However there is one important difference which is key to obtaining a uniform bound on the excess loss of our monotone algorithm. Indeed, Pestov does not regularize by adding noise which requires him to refine the partition element under some very restrictive conditions (the conditional loss on the partition should not be close to $1/2$ nor to $0$ or~$1$) which has the effect of requiring to make a very large increases of the sample size between two stages, resulting in a slower, non-uniform, convergence rate (more specifically, he does not provide an explicit formula for computing~$N$ from~$n$).

\section{General Framework}\label{sec:general}
Given an algorithm $A$ which maps an input sample $S$ to an output hypothesis $A(S)$, 
    we construct a monotone algorithm $M$ using two intermediate algorithms:
\begin{enumerate}
    \item A \emph{regularization} algorithm $R$ is an algorithm that takes as input 
        a sample $S$ and a hypothesis~$h$ and returns a (possibly randomized) hypothesis $R(h,S)$. 
        It might be useful/intuitive to think about $R(h,S)$ as a smooth/regularized version of $h$.
    \item
    An \emph{update} algorithm $U$ is an algorithm that takes as an input a sample~$S$ 
        and two hypotheses: (i) $h_0$ which is called the \emph{current} hypothesis, 
        and (ii) $h_1$ which is called the \emph{candidate} hypothesis. 
        The algorithm then outputs a hypothesis denoted by $U(h_0,h_1,S)\in\{h_0,h_1\}$. 
        Intuitively,~$U$ chooses whether to replace the current hypothesis $h_0$ 
        with the candidate hypothesis $h_1$, when the latter has smaller error.
\end{enumerate}
The monotone algorithm will then be constructed in an iterative manner from $A$ by
    applying $A$ to prefixes of the training sample of increasing size 
    and using the \emph{update} algorithm at each step to decide whether 
    to keep the current hypothesis or to update it to the new one (built on a longer prefix). 
    At the end, we output a normalized version of the currently chosen hypothesis.

The \emph{update} algorithm ensures that with high probability we update the hypothesis 
    only when the new hypothesis has better (smaller) loss than the previous one. 
    But since there is still a small chance we have updated to a worse hypothesis, 
    the regularization step will be used to correct for corresponding additional expected loss.

See Figure~\ref{fig:algM} for a more precise description of the algorithm $M$.
\begin{figure}
\begin{tcolorbox}
\begin{center}
{\bf The Monotonizing Algorithm $M$}
\end{center}
\begin{itemize}
    \item Input: a learning algorithm $A$, a regularization algorithm $R$, an update algorithm $U$, an increasing function $b:\mathbb{N}\rightarrow \mathbb{N}$ such that $b(0)=1$, 
          and a sample $S\sim D^m$ of iid examples from $D$.
    \item Output: a hypothesis determined as follows
\begin{enumerate}
    \item If $m< 2\cdot b(1) + b(0)$ then output $R(f,Z)$, where $f=A(\emptyset)$
    and $Z$ consists of the first example from $S$.
    \item Else, let $T\geq 2$ be maximal such that $b(T-1)+\sum_{t=0}^{T-1} b(t)\le m$; discard from $S$ the last $m - \bigl(b(T-1)+\sum_{t=0}^{T-1}b(t)\bigr)$ examples.
    \item Partition the remaining examples into $T+1$ blocks $\{B_t\}_{t=0}^{T}$ where $\lvert B_t\rvert=b(t)$ for $t\leq T-1$ and $\lvert B_T\rvert = b(T-1)$.
    \item Let $S_t$ denote the union of the first $t$ blocks: $S_t := \bigcup_{i=0}^t B_i$. 
        \item Set $f_{0}:=A(\emptyset)$.
    \item For each  $t=1,\ldots,T-1$ perform the following operations:
    \begin{enumerate}
        \item Compute the new candidate hypothesis using $A$: $h_{t}:=A(S_{t-1})$.
        \item Choose the new hypothesis between $f_{t-1},h_t$: $f_{t}:=U(f_{t-1}, h_{t}, B_{t})$.
    \end{enumerate}
    \item On the last block, output $R(f_{T-1},B_{T})$.
\end{enumerate}
\end{itemize}
\end{tcolorbox}
\caption{\label{fig:algM}
Pseudo-code for the transformation $M$ which converts any learning rule $A$ to a monotone one with similar guarantees. The algorithm proceeds by running $A$ on increasing prefixes of the input sample and apply carefully tailored model selection from the outputs of $A$.
Item 1 handles a trivial base-case and can be ignored at first read.
Note that the last two blocks have identical size $\lvert B_{T-1}\rvert = \lvert B_T\rvert =b(T-1)$. 
This is because the  last block serves for regularization, while the first $T-1$ blocks are used for training and model selection.
{Also notice that the regularizer is applied only on the last iteration, and not on each iteration. 
This is because the noise is not needed when one compares different hypotheses. 
Its role is to slightly deteriorate the loss of the output in order to compensate for future mistakes.}
}
\end{figure}

\paragraph{A Framework for Proving Monotonicity.}
We  introduce several conditions on the update and regularization algorithms 
    which guarantee the success of  Algorithm $M$ when applied to \emph{any} learning algorithm $A$.
    We then analyze our algorithms by showing that they satisfy these conditions.
    Let us begin by introducing some notation: 
    given a hypothesis $h$, denote by $\R_n(h)$ the quantity
    \begin{equation}\label{eq:rnn}
    \R_n(h):= \E_{S\sim D^{n}}\Err R(h,S)        
    \end{equation}
    where the expectation is with respect to the sample $S$ of size $n$.
\begin{definition}[Sufficient Conditions for Monotonicity]
\label{def:suff-mon}
    Let $R$ be a regularization algorithm, let~$U$ be an update algorithm. We say that \emph{$(R,U)$ are successful} if for every source distribution $D$ the following conditions are satisfied: 
\begin{enumerate}[{\bf(C1)}]
    \item After regularization, the expected loss of the update algorithm is \emph{non-increasing}:
    \[
    (\forall n\in\mathbb{N}) (\exists N\ge n) (\forall h_0,h_1): \E_{S\sim D^N}\R_N(U(h_0,h_1,S)) \le \R_n(h_0).
    \]
    For $n\in\mathbb{N}$, let $N(n)$ denote the smallest number $N>n$ for which the above holds.
    \item The \emph{update} algorithm competes with the new hypothesis at a small cost: 
    \[
    (\forall n\in\mathbb{N})(\forall h_0,h_1): \E_{S\sim D^n}\R_n(U(h_0,h_1,S)) \le \Err(h_1)+\cost(n),
    \]
    for some function $\cost$ such that $\lim_{n\to\infty} \cost(n)=0$.
\end{enumerate}
\end{definition}

\begin{prop}[Monotonicity and Competitiveness of $M$]\label{c:final}
Let $R,U$ be a regularization and update algorithm such that $(R,U)$ are successful. 
    Define $b:\mathbb{N}\to\mathbb{N}$ according to the recurrence $b(0)=1$ and $b(t+1)=N(b(t))$, where $N(\cdot)$ is the function defined in Condition (C1) of Definition \ref{def:suff-mon}.
    Consider  algorithm $M$ with $(R,U)$ and $b$ as inputs.
    Then, for every algorithm $A$, applying $M$ on $A$ yields a monotone algorithm 
    whose performance is competitive with that of $A$:
    for every source distribution $D$ and every sample size $m\geq 2b(1) + b(0)$,
\[
\E_{S\sim D^m}\Err(M(S)) \le \E_{S\sim D^m}\Err(A(S_{T-2})) + \cost(b(T-1)),
\]
where $T$ and $S_{T-2}$ are as in the pseudo-code of $M$ in Figure~\ref{fig:algM}. 
\end{prop}
\begin{proof}
The case of $m < 2b(1) + b(0)$ is trivial, 
    so we assume that $m \geq 2b(1) + b(0)$.
    We start by proving  monotonicity. 
    We only need to consider the case where a new block is added 
    (otherwise, the additional examples are simply discarded and thus the expected error is unchanged). 
    In this case, it is sufficient to consider the effect of adding a new block $B_{T}$ and prove that
    \begin{equation}\label{eq:monsufficient}
     \E \R_N(f_{T-1})\leq \E \R_n(f_{T-2}),   
    \end{equation}
    for $n=b(T-2)$ and $N=N(n)=b(T-1)$.
    
Conditioned on $S_{T-2}$, the hypotheses $f_{T-2}$ and~$h_{T-1}$ are fixed and 
    \[f_{T-1}=U(f_{T-2},h_{T-1},B_{T-1})\] 
    is a function of $B_{T-1}$ (which is not under conditioning).
    Therefore, 
\begin{align*}
    \E\bigl[\R_N(f_{T-1}) \big\vert S_{T-2}\bigr]
    &=
    \E_{B_{T-1}\sim D^N}\bigl[\R_N(U(f_{T-2},h_{T-1},B_{T-1}))\big\vert S_{T-2}\bigr]\\
    &\leq
    \E\bigl[\R_n(f_{T-2})\big\vert S_{T-2}\bigr], 
\end{align*}
{where the last inequality follows by of Condition (C1)\footnote{{Notice that we apply (C1) here while conditioning on $S_{T-2}$. This is valid because $B_{T-1}$ is independent of $S_{T-2}$ and therefore Condition (C1) applies for every fixing of $S_{T-2}$.}}.}
Equation~(\ref{eq:monsufficient}) now follows by taking expectation over $S_{T-2}$.

For the second part, observe that Condition (C2) implies that
\begin{align*}
\E_{S\sim D^m}\Err(M(S)) &= \E\Err R(f_{T-1},B_{T}) \\
 &= \E \R_N\bigl(U(f_{T-2},h_{T-1},B_{T-1})\bigr)\\
&\le  \E\Err(h_{T-1}) + \cost(N) \tag{Condition (C2)}\\
&= \E\Err(A(S_{T-2})) + \cost(b(T-1)). 
\end{align*}
\end{proof}

\subsection{The Base-Class Approach}
We design two algorithms using our  framework above, 
    one in binary classification which serves as a ``warmup'' 
    and a more general one in multiclass classification.
    In this subsection we describe a common abstraction of these two algorithms.
    We hope this abstraction will be useful in other contexts as well. (E.g., other loss functions.)

The common abstraction boils down to assuming that every hypothesis $h$ 
    has an associated \emph{simple} hypothesis class $B_h$ such that $h\in B_h$.
    For example in the context of binary classification ($Y=\{0,1\}$) our $B_h$ will consist of two hypotheses: $h$ and its negation $1-h$; i.e., $B_h=\{h, 1-h\}$. 
    More generally in multiclass classification with $k$ labels our $B_h$ 
    will consist of $k$ hypotheses.

We require that $B_h$ is ``well-behaved'' in a precise sense which we next describe.
    Consider the following randomized empirical risk minimizer over $B_h$.
\begin{tcolorbox}
\begin{center}
{\bf Algorithm $G$: Randomized Empirical Risk Minimization}
\end{center}
\begin{itemize}
    \item Input: a hypothesis $h\in Y^X$ and a sample $S\in (X\times Y)^n$
    \item Set $B_h^\star = \{f\in B_h : \EmpS(f)=\min_{g\in B_h}\EmpS(g)\}$
    \item Output: a uniformly random hypothesis from $B_h^\star$, which is denoted by $G(h,S)$.
\end{itemize}
In words, $G(h,S)$ is a random empirical risk minimizer in $B_h$.
\end{tcolorbox}    

Note that on the empty sample, $G(h,\emptyset)$ is a random hypothesis drawn uniformly from $B_h$.

Let $\G_n(h)$ denote the expected loss of $G(h,S)$ where  $S$ is of size $n$:
\[
\G_n(h) := \E_{S\sim D^n} \Err(G(h,S))\,.
\]

\begin{definition}[Successful Base-Class]\label{def:base-class}
Let $h\mapsto B_h$ be a mapping which associates with every hypothesis $h$
a finite hypothesis class $B_h$ such that $h\in B_h$.
This mapping is called \emph{successful} if the following properties are satisfied:
\begin{enumerate}
    \item The loss of the randomized ERM over $B_h$ is monotone
    \[
    (\forall D)(\forall h)(\forall n): \G_{n+1}(h)\le \G_n(h)\,.
    \]
    \item There exists a function $\unif(n)$ with $\lim_{n\to\infty} \unif(n)=0$ such that
    every $B_h$ satisfies uniform convergence with rate $\unif(n)$:
    \[
    (\forall D)(\forall h)(\forall n): 
    \E_{S\sim D^n}\Bigl[\max_{f\in B_h}\bigl\lvert\EmpS(f) - \Err(f) \bigr\rvert\Bigr] \leq  \unif(n).
    \]
    \item The loss of the random ERM on the empty sample is independent of $h$ and of the source distribution $D$:
    \[
    (\forall D)(\exists \alpha)(\forall h): \G_0(h)=\alpha\,.
    \]
    In other words, the expected loss of a uniform random hypothesis from $B_h$
    is equal to a universal constant $\alpha$ which does not depend on $h$. ($\alpha$ can depend on the source distribution~$D$.)
\end{enumerate}
\end{definition}
\begin{remark}
The first condition above could be relaxed into $\G_{N}(h)\le \G_n(h)$ for some $N\ge n$ large enough. Indeed this would be sufficient to derive condition (C1). But we will show that the randomized ERM that we consider actually satisfies the stronger property of monotonicity for $N=n+1$.
\end{remark}
\begin{remark}
The second item above implies (via a triangle inequality) that the randomized ERM $G(h,\cdot)$ is competitive with~$h$: 
\begin{equation}\label{eq:ERMcomp}
    (\forall D)(\forall h)(\forall n): 0\leq  \G_n(h)-\min_{f\in B_h}\Err(f) \leq 2\unif(n).
\end{equation}
(Note that $\min_{f\in B_h}\Err(f)\leq \Err(h)$, since $h\in B_h$.)
\end{remark}

\subsubsection*{The Regularization and Update Algorithms}
We next describe how a successful map $h\mapsto B_h$ 
    yields a successful pair of regularization and update algorithms.

\begin{tcolorbox}
\begin{center}
{\bf The Regularization Algorithm $R$}
\end{center}
\begin{itemize}
    \item Input: a hypothesis $h\in Y^X$, a sample $S\in (X\times Y)^n$.
    \item Output: with probability $1-\eta_n$ output $G(h,S)$
    and with probability $\eta_n$ output $G(h,\emptyset)$.
    Here $\eta_n\in (0,1)$ is a decreasing function of $n$ satisfying $\lim_{n\to\infty}\eta_n=0$.
\end{itemize}
\end{tcolorbox}

\begin{tcolorbox}
\begin{center}
{\bf The Update Algorithm $U$}
\end{center}
\begin{itemize}
    \item Input: two hypotheses $h_0,h_1\in Y^X$, a sample $S\in (X\times Y)^{n}$. 
    \item Compute $\min_{f\in B_{h_0}}\EmpS(f)$ and $\min_{f\in B_{h_1}}\EmpS(f)+ \eps_n$,
    and output $h_0$ if the first quantity is smaller and $h_1$ otherwise.
    Here $\eps_n\in (0,1)$ is a decreasing function of $n$ satisfying $\lim_{n\to\infty}\eps_n=0$.
\end{itemize}
\end{tcolorbox}

\begin{prop}[Base-class]\label{prop:base-class}
Assume $h\mapsto B_h$ is successful with uniform convergence rate~$\unif(n)$.
Then, the update algorithm $U$ and the regularization algorithm $R$
with parameters 
\[\eta_n = \frac{1}{2\sqrt{n}}\text{ and } \new{\eps_n=\sqrt{{\ln(64 n)}/{n}}+6\unif(n)}\]
satisfy Condition (C1) with $N(n) = 4n$ and condition (C2) with $\cost(n) = 2\eta_n+3\eps_n$.
\end{prop}

The rest of this section is dedicated to proving Proposition~\ref{prop:base-class}.
    We begin with collecting some simple properties 
    of the regularization and update algorithms, 
    and then continue to establish Conditions (C1) and (C2).

\subsection{Basic Properties of $R$ and $U$}
Fix a source distribution $D$.
Recall the definition of $\R_n$ (Equation~\ref{eq:rnn}):
\[\R_n(h)= \E_{S\sim D^{n}}\Err R(h,S)\]
We next present some basic properties of $\R_n$ which will be useful in proving Proposition~\ref{prop:base-class}.

A simple calculation yields the following relationship between $\G_n$ and $\R_n$: 
for every hypothesis~$h$ and every $n$:
\begin{align}
\R_n(h) &= (1-\eta_n)\cdot \G_n(h) + \eta_n\cdot \G_0(h) \notag\\
        &=(1-\eta_n)\cdot \G_n(h) + \eta_n\cdot \alpha\label{e:ddefnoise}\\ 
        &= \G_n(h) + \eta_n\left(\alpha - \G_n(h) \right)\label{e:ddefnoise2}.
\end{align}

The following claim asserts that $\R_n$ is monotone:
\begin{claim}\label{c:rmonn}
For all $n,m$, if $m\ge n$ and $\eta_m\le \eta_n\le 1$ then 
\[
\R_n(h)-\R_m(h) \ge (\eta_n-\eta_m)\left(\alpha-\G_m(h)\right)\,,
\]
and in particular,
\[
\forall h,\,\R_m(h) \le \R_n(h)\,.
\]
\end{claim}
\begin{proof}
By Equation \eqref{e:ddefnoise2}:
\begin{align}
\R_n(h) - \R_m(h) &= (1-\eta_n)\left(\G_n(h) - \G_m(h)\right) + (\eta_n-\eta_m)\left(\alpha-\G_m(h)\right)\notag\\
&\ge (\eta_n-\eta_m)\left(\alpha-\G_m(h)\right)
\tag{$\G_m(h) \le \G_n(h)$ by Definition \ref{def:base-class}}\\
&\ge 0 \tag{$\G_m(h)\le \G_0(h) = \alpha$ by Definition \ref{def:base-class}}
\end{align}
which proves the claim.
\end{proof}

Lastly, observe that $R$ does not deteriorate the performance of $h$:
\begin{claim}\label{c:Rcomp}
For all $n$,
\[
\R_n(h) \le \Err(h)+\eta_n + 2\unif(n)\,.
\]
\end{claim}
\begin{proof}
\begin{align*}
\R_n(h) &= \G_n(h) + \eta_n\left(\alpha - \G_n(h) \right) \tag{Equation~\eqref{e:ddefnoise2}}\\
        &\leq \G_n(h) + \eta_n \tag{$\G_n(h)\le \G_0(h)=\alpha \le1$}\\
        &\leq \Err(h)+\eta_n  + 2\unif(n). \tag{Equation~\eqref{eq:ERMcomp}}
\end{align*}
\end{proof}

\subsection{Update Probability}
We first provide upper and lower bounds on the probability of making an update.
\begin{lemma}\label{le:prob}
Let $\new{u_n=\eps_n-6\unif(n)} = \sqrt{\ln(64n)/n}$. (Recall the definition of $\eps_n$ in Proposition~\ref{prop:base-class}.)
If $\R_n(h_1)>\R_n(h_0)$ then
\[
\Pr_{S\sim D^n}[U(h_0,h_1,S)=h_1] \le 4\exp\bigl({-nu_n^2/2}\bigr) = \frac{1}{2\sqrt{n}} = \eta_n\,,
\]
and if $\R_n(h_1)<\R_n(h_0)-2\eps_n$ then
\[
\Pr_{S\sim D^n}[U(h_0,h_1,S)=h_1] \ge 1-4\exp\bigl({-nu_n^2/2}\bigr)\,.
\]
\end{lemma}
\begin{proof} 
To simplify the calculations below, for a hypothesis $h$ let 
\begin{align}
 \ErrS(h)&:=\min_{f\in B_{h}}\EmpS(f)\label{eq:lstars}\\
 \Errs(h)&:=\min_{f\in B_{h}}\Err(f)\label{eq:lstard}
\end{align}
We first connect the performance of the regularized versions of the hypotheses $h_0$ and $h_1$ to their error probability $\Errs(h_0)$ and $\Errs(h_1)$.
From Equation~\eqref{e:ddefnoise} we obtain
\begin{equation}\label{e:rg}
\R_n(h_1)-\R_n(h_0) = (1-\eta_n)(\G_n(h_1)-\G_n(h_0))\,.
\end{equation}
Recall from the pseudo-code of the update algorithm $U$ that the probability of update, which we denote $p_n$ is given by
\[
p_n:=\Pr_{S\sim D^n}[U(h_0,h_1,S)=h_1]=\Pr_{S\sim D^n}\Bigl[\ErrS(h_0)-\ErrS(h_1) >\epsilon_n \Bigr].
\]
Hence we have the following implications:
\begin{align}
    \R_n(h_1)>\R_n(h_0) &\Rightarrow \G_n(h_1) > \G_n(h_0) \tag{by Equation~\eqref{e:rg}}\\
    &\Rightarrow \Errs(h_1) + 2\unif(n) > \Errs(h_0) \tag{by Equation~\eqref{eq:ERMcomp}}\\
    &\Rightarrow p_n \le \Pr\left[\lvert\ErrS(h_0)-\Errs(h_0)\rvert+\lvert\ErrS(h_1)-\Errs(h_1)\rvert>\epsilon_n - 2\unif(n)\right]\notag
\end{align}
and similarly
\begin{align}
    \R_n(h_0)>\R_n(h_1) + 2\eps_n &\Rightarrow \G_n(h_0) > \G_n(h_1) + 2\eps_n\tag{by Equation~\eqref{e:rg} and $\eta_n<1$}\\
    &\Rightarrow \Errs(h_0) + 2\unif(n) > \Errs(h_1) + 2\eps_n \tag{by Equation~\eqref{eq:ERMcomp}}\\
    &\Rightarrow 1-p_n \le \Pr\left[\lvert\ErrS(h_0)-\Errs(h_0)\rvert+\lvert\ErrS(h_1)-\Errs(h_1)\rvert>\epsilon_n - 2\unif(n)\right]\notag
\end{align}
We have in both cases to upper bound
\[
\Pr\left[\lvert\ErrS(h_0)-\Errs(h_0)\rvert+\lvert\ErrS(h_1)-\Errs(h_1)\rvert> \epsilon_n-2\unif(n)\right]\,,
\]
which is less than
\[
\Pr\left[\lvert\ErrS(h_0)-\Errs(h_0)\rvert>\epsilon_n/2 - \unif(n) \right] +\Pr\left[\lvert\ErrS(h_1)-\Errs(h_1)\rvert>\epsilon_n/2 - \unif(n) \right]
\]
We can apply Mcdiarmid's inequality~\citep*{Mcdiarmid} to both terms. Recall that (a special case of) Mcdiarmid's inequality asserts that $\Pr[\lvert Z-\E Z\rvert \geq \eps]\leq2\exp(-2n\eps^2)$ for every random function $Z=Z(V_1,\ldots, V_n)$, where the $V_i$'s are independent,
and such that $Z$ is stable in the sense that $\lvert Z(\vec v') - Z(\vec v'')\rvert \leq 1/n$ for every
pair of vectors $\vec v', \vec v''$ whose hamming distance is $1$. Here it is applied on $Z(S)=\ErrS(h)=\min_{f\in B_{h}}\EmpS(f)$.
\new{Let $h\in\{h_0,h_1\}$,}
\begin{align*}
\Pr\left[\bigl\lvert\ErrS(h)-\Errs(h)\bigr\rvert>\epsilon_n/2 - \unif(n) \right] &=
\Pr\left[\bigl\lvert\min_{f\in B_{h}}\EmpS(f)- \min_{f\in B_{h}}\Err(f)\bigr\rvert>\epsilon_n/2 - \unif(n)\right]
\end{align*}
\new{Note that}
\begin{align*}\bigl\lvert\min_{f\in B_{h}}\EmpS(f)- \min_{f\in B_{h}}\Err(f)\bigr\rvert &\leq  \lvert\min_{f\in B_{h}}\EmpS(f)- \E\min_{f\in B_{h}}\EmpS(f)\rvert + \lvert\E\min_{f\in B_{h}}\EmpS(f) - \min_{f\in B_{h}}\Err(f)\rvert\\
&\leq 
\lvert\min_{f\in B_{h}}\EmpS(f)- \E\min_{f\in B_{h}}\EmpS(f)\rvert + 2\unif(n)
\tag{$\lvert\E\min_{f\in B_{h}}\EmpS(f) - \min_{f\in B_{h}}\Err(f)\rvert \leq 2\unif(n)$}
\end{align*}
\new{Therefore,}
\begin{align*}
\Pr\left[\bigl\lvert\ErrS(h)-\Errs(h)\bigr\rvert>\epsilon_n/2 - \unif(n) \right]&\leq 
\Pr\left[\lvert\min_{f\in B_{h}}\EmpS(f)- \E\min_{f\in B_{h}}\EmpS(f)\rvert +  2\unif(n) >\epsilon_n/2 - \unif(n) \right]\\
&=
\Pr\left[\lvert\min_{f\in B_{h}}\EmpS(f)- \E\min_{f\in B_{h}}\EmpS(f)\rvert >u_n/2 \right]
\tag{$u_n= \epsilon_n - 6\unif(n)$}\\
&\leq 2\exp\bigl({-2n(u_n/2)^2}\bigr)
\tag{\new{By Mcdiarmid's inequality}}
\end{align*}
We thus get the following upper bound:
\[
2\cdot 2\exp\bigl({-2n(u_n/2)^2}\bigr) = 4\exp\bigl({-nu_n^2/2}\bigr)\,.
\]
\end{proof}

\subsection{Condition (C1)} 

\begin{lemma}\label{lem:c1parameters}
Under the assumptions of Proposition~\ref{prop:base-class}, Condition (C1) is satisfied with $N(n)=4n$.
\end{lemma}
\begin{proof} 
Let $n\in\mathbb{N}$ and $N(n)=4n$.
Set $p_N:=\Pr_{S\sim D^N}[U(h_0,h_1,S)=h_1]$, the left-hand side of (C1) can be written as
\begin{align}
\E_{S\sim D^N}\R_N(U(h_0,h_1,S)) &= p_N\R_N(h_1)+(1-p_N)\R_N(h_0) \label{eq:cnn}
\end{align}
Assume first that $\R_N(h_1)\le \R_n(h_0)$. 
    In this case, since $N>n$, we have $\R_N(h_0)\le \R_n(h_0)$.
    Thus, Equation~\eqref{eq:cnn} shows that $\E_{S\sim D^N}\R_N(U(h_0,h_1,S))$ is a convex combination 
    of two terms both $\leq \R_n(h_0)$ hence $\E_{S\sim D^N}\R_N(U(h_0,h_1,S))\le \R_n(h_0)$.

Thus, assume that $\R_N(h_1)>\R_n(h_0)$. 
By Equation~\eqref{eq:cnn},
 \[
 \E_{S\sim D^N}\Bigl[\R_N(U(h_0,h_1,S))-\R_n(h_0)\Bigr]=p_N\left(\R_N(h_1)-\R_N(h_0)\right)+\R_N(h_0)-\R_n(h_0)\,.
 \]
Therefore, it suffices to show that in this case
\[
p_N\bigl(\R_N(h_1)-\R_N(h_0)\bigr)\le\R_n(h_0)-\R_N(h_0).
\]
Denoting $\eta'=\eta_n-\eta_N\ge 0$, we have, by Claim \ref{c:rmonn}:
\begin{equation}
\R_N(h_0)-\R_n(h_0) \le -\eta'\bigl(\alpha-\G_N(h_0)\bigr)
\end{equation}
Further, because $\G_N(\cdot)\leq \alpha$,
\[
\R_N(h_1)-\R_N(h_0) = (1-\eta_N)\left(\G_N(h_1)-\G_N(h_0)\right) \le \alpha-\G_N(h_0).
\]
The last two inequalities yield:
\begin{align*}
    p_N\bigl(\R_N(h_1)-\R_N(h_0)\bigr)&\le p_N\Bigl(\alpha-\G_N(h_0)\Bigr)\\
    \eta'\left(\alpha-\G_N(h_0))\right) &\le \R_n(h_0)-\R_N(h_0).
\end{align*}
Hence, condition (C1) is satisfied provided that $p_N \le \eta'$. 
By Lemma~\ref{le:prob}, we have
\[
p_N\le 2\cdot 2\exp\bigl({-2N(u_N/2)^2}\bigr) = 4\exp\bigl({-Nu_N^2/2}\bigr) = 4\exp\Bigl({-\frac{1}{2}\ln (64 N)}\Bigr) = \frac{1}{2\sqrt{N}}\,.
\]
Now we can verify that for $N=4n$,
\[\eta'=\frac{1}{2\sqrt{n}}-\frac{1}{2\sqrt{N}}= \frac{1}{2\sqrt{N}},\]
and hence
\[
p_N\le \eta'
\]
which concludes the proof.
\end{proof}

\subsection{Condition (C2)}
\begin{lemma}\label{lem:parameters}
Under the assumptions of Propositon~\ref{prop:base-class}, Condition (C2) is satisfied with
\[
\cost(n)=2\eta_n+3\eps_n\,.
\]
\end{lemma}
\begin{proof}
Let $q_n:=\Pr_{S\sim D^n}[U(h_0,h_1,S)=h_0]$ be the probability of not switching to $h_1$.
    Thus,
\begin{align*}
\E_{S\sim D^n}\R_n(U(h_0,h_1,S)) &= q_n\R_n(h_0) + (1-q_n)\R_n(h_1)
\end{align*}
Therefore, if $\R_n(h_0)\leq \R_n(h_1) + 2\eps_n$ then : 
\begin{align*}
\E_{S\sim D^n}\R_n(U(h_0,h_1,S))&\leq \R_n(h_1) + 2\eps_n\\
                                &\leq \Err(h_1) +  \eta_n + 2\unif(n) + 2\eps_n \tag{by Claim~\ref{c:Rcomp}}      \\
                                &\leq \Err(h_1) + 2\eta_n + 3\eps_n \tag{$2\unif(n)\leq \eps_n$}\,.
\end{align*}
Thus, the conclusion holds in this case.
Therefore, assume that $\R_n(h_0)> \R_n(h_1) + 2\eps_n$. In this case,
\begin{align*}
\E_{S\sim D^n}\R_n(U(h_0,h_1,S)) &= q_n\R_n(h_0) + (1-q_n)\R_n(h_1)\\
                                 &\leq q_n + \R_n(h_1). \tag{$q_n,\R_n(h_0)\in [0,1]$}\\
                                 &\leq \R_n(h_1) + \eta_n \tag{by Lemma~\ref{le:prob}}\\
                                &\leq \Err(h_1) +  2\eta_n + 2\unif(n)\tag{by Claim~\ref{c:Rcomp}}       \\
                                &\leq \Err(h_1) + 2\eta_n + \eps_n \tag{$2\unif(n)\leq \eps_n$}\,.
\end{align*}

\end{proof}

\section{Binary Classification}\label{sec:binary}

In this section we prove that in the setting of binary classification,
    one can associate with every hypothesis $h$ a base class $B_h$ 
    such that Definition~\ref{def:base-class} is satisfied.
\begin{proposition}[Successful Base-Class: Binary Classification]
Let the label-space $Y$ be $Y=\{0,1\}$. For each $h:X\to Y$ let $B_h=\{h, 1-h\}$.
Then, the mapping $h\to B_h$ is successful with uniform convergence rate $\unif(n) = 1/\sqrt{n}$ and with $\alpha=1/2$.
\end{proposition}
\begin{proof}
That $\unif(n)=1/\sqrt{n}$ follows by elementary probabilistic argument (essentially bounding the variance of a Binomial random variable):
\begin{align}
    \unif(n) &= \E_{S\sim D^n}\max\left(\left|\EmpS(h)-\Err(h)\right|, \left|\EmpS(1-h)-\Err(1-h)\right|\right)\notag\\
    &= \E_{S\sim D^n}\left|\EmpS(h)-\Err(h)\right|\tag{\new{$\EmpS(h)+\EmpS(1-h)=\Err(h)+\Err(1-h)=1$}}\\
    &\le \sqrt{\E_{S\sim D^n}\left(\EmpS(h)-\Err(h)\right)^2}\tag{By Jensen's inequality}\\
    &= \sqrt{\mathtt{Var}(\EmpS(h))}\notag\\
    &\le \frac{1}{\sqrt{n}}.\notag
\end{align}

Also, showing that $\alpha=1/2$ is simple:
indeed for every distribution~$D$:
\begin{align*}
\G_{0}(h) = \frac{1}{2}\Err(h) + \frac{1}{2}\Err(1-h) = \frac{1}{2}\Err(h) + \frac{1}{2}\bigl(1-\Err(h)\bigr) = \frac{1}{2}.
\end{align*}

Thus,  it remains to prove the first Item in Definition~\ref{def:base-class}:
\begin{lemma}\label{c:gmon}
For all $n,m$ such that $m\ge n$ we have 
\[
\forall h,\,\G_m(h) \le \G_n(h).
\]
\end{lemma}
We note that \cite*{pestov2021universally} proved this statement when $m,n$ are odd.
(See Lemma 3.1 in \citep*{pestov2021universally}).
We defer the proof to the next section where we derive a more general result that
applies to multiclass with an arbitrary number of labels $k$. (See Lemma~\ref{le:monmulti}.)
\end{proof}

\section{Multiclass Classification}\label{sec:multiclass}

We consider the multiclass case with $k<\infty$ labels. 
A hypothesis is a map from $X$ to $[k]$. 
To define $B_h$ we introduce cyclic permutations $s_0,\ldots,s_{k-1}$ of $[k]$ ($s_i(j)=(j+i) \mod k$) and let $B_h=\{ s_i\circ h : i\leq k\}$.
\begin{proposition}[Successful Base-Class: Multiclass Classification]\label{prop:multiclass}
Let the label-space $Y$ be $Y=[k]$. For each $h:X\to Y$ let $B_h=\{ s_i\circ h : i\leq k\}$.
Then, the mapping $h\to B_h$ is successful with uniform convergence rate $\unif(n) = 36/\sqrt{n}$ and with $\alpha=\frac{k-1}{k}$.
\end{proposition}
\begin{proof}
Let $h\in H$. We begin with the simplest part: namely that $\alpha = \frac{k-1}{k}$. 
    Notice that the events
\[{E}_i = \{(x,y) : s_i \circ h (x) = y\}\]
    are pairwise disjoint; 
    in fact they form a partition of $X\times Y$ because for each $(x,y)$
    there is a unique~$i$ such that $ s_i\circ h (x) = y$).
    Therefore, for every distribution $D$ over $X\times Y$:
\begin{align*}
\G_{0}(h) &= \frac{1}{k}\sum_{i=1}^k\Err( s_i\circ h)\\
          &= \frac{1}{k}\sum_{i=1}^k\bigl(1- D({E}_i)\bigr)\\
          &= \frac{1}{k}\bigl(k - \sum_{i=1}^k D({E}_i)\bigr)\\
          &=\frac{k-1}{k}.
\end{align*}

To see that $\unif(n)=36/\sqrt{n}$ notice that the family of events $\mathcal{E} = \{E_i : i\leq k\}$
consists of pairwise disjoint events and therefore its VC dimension is $1$.
By VC theory (see\footnote{Theorem 1.16 in \citep*{lugosi2002pattern} gives $\E\sup_{E\in\mathcal{E}} \left\lvert D_S(E)-D(E)\right\rvert\le \frac{24}{\sqrt{n}} \int_0^1 \sqrt{\ln (2N(\epsilon))} d\epsilon$, where $N(\eps)$ denotes the covering number of the family $\mathcal{E}$. Here, since the $E_i$'s are disjoint, $N(\epsilon)\le 1+1/\epsilon$, and we have $\int_0^1 \sqrt{\ln (2+2/\epsilon)} d\epsilon \approx 1.22 \le \frac{3}{2}$ so we get $\E\max_{i} \left\lvert D_S(E_i)-D(E_i)\right\rvert\le \frac{36}{\sqrt{n}}$} e.g.,~\citep*{lugosi2002pattern})
\begin{align*}
\frac{36}{\sqrt{n}}&\geq \E_{S\sim D^n}\max_i\bigl\lvert D_S(E_i) - D(E_i)\bigr\rvert\\
                 &= \E_{S\sim D^n}\max_i\bigl\lvert \bigl(1-D_S(E_i)\bigr) - \bigl(1-D(E_i)\bigr)\bigr\rvert\\
                 &= \E_{S\sim D^n}\max_i\bigl\lvert \EmpS(s_i\circ h) - \Err(s_i\circ h )\bigr\rvert.
\end{align*}
Above, $D_S$ denotes the empirical distribution induced by the sample $S=\{(x_i,y_i)\}_{i=1}^m$
(i.e., $D_S(E) = \frac{1}{m}\sum_{i=1}^{m}1[(x_i,y_i)\in E]$.).
Thus, it remains to prove the first Item in Definition~\ref{def:base-class}:

\begin{lemma}\label{le:monmulti}
For all $n,m$ such that $m\ge n$ we have 
\[
\forall h,\,\G_m(h) \le \G_n(h).
\]
\end{lemma}
\begin{proof}
By induction, it suffices to consider the case of $m=n+1$.

We first reformulate this problem in simpler terms. 
    Recall that any example $z=(x,y)$ is classified correctly by exactly one of the $h_j$'s in $B_h$. 
    Thus, partition the domain into sets $E_1,\ldots,E_k$ (with $k=\lvert B_h\rvert$) 
    such that $z\in E_j\Leftrightarrow h_j(x)=y$. 
    Thus, we have $\Pr(E_j)=1-\Err(h_j)$.
    To simplify the expressions below we denote
\[
p_i=1-\Err(h_i),\,\, q_i=\Err(h_i)\,.
\]

Let $S=\{z_i\}_{i=1}^n\sim D^n$ be an i.i.d sample and let $i\leq n$.
    Define $X_i$ to be the random variable which is equal 
    to the unique index $j$ such that $z_i\in E_j$. 
    (I.e., $h_j$ correctly classifies $z_i$.)
    Notice that the variables $(X_1,\ldots,X_n)$ are i.i.d.\ with values in $[k]$ 
    and distribution given by $\Pr(X_i=j)=p_j$ for all $i\leq n,j\leq k$.
    We can then express $\G_n(h)$ as the expectation over the sample ${L_n}=(X_1,\ldots,X_n)$ of the quantity 
\[
f({L_n}):=\frac{1}{\lvert I \rvert}\sum_{j\in I}q_j,\,
\] 
where $I=I({L_n})$ is the set of indices $j$ such that $\sum_{i=1}^n 1[X_i=j]$ is largest.
(Equivalently, such that $h_j$ is an empirical risk minimizer in $B_h$ with respect to the input sample $S$.)

Let ${L=L_{n+1}}$ be a sample of $n+1$ such variables $(X_1,\ldots,X_{n+1})$ (corresponding to an input sample $S\sim D^{n+1}$), and let $L^{-i}$ denote the sample $L$ without its $i$-th element, by symmetry and by linearity of expectation, we have
\[
\G_{n+1}(h)-\G_n(h) = \E_{L} \left[f(L)-f(L^{-(n+1)})\right] =  \frac{1}{n+1}\sum_{i=1}^{n+1}\E_{L} \left[f(L)-f(L^{-i})\right]
\]
We first study the above quantity when conditioned on $\lvert I\rvert>1$ 
(i.e., there are at least $2$ empirical risk minimizers in $B_h$).
Notice that when $\lvert I\rvert > 1$ every $i\leq n+1$ satisfies $I_{i}\subseteq I$,
where $I_{i}=I(L^{-i})$. Also notice that {if $\lvert I\rvert > 1$ then} $I_{i}=I$ if and only if $X_i\notin I$. Thus,
\begin{align}
&\sum_{i=1}^{n+1}\E_{L} \left[\left.f(L)-f(L^{-i}) \right| |I|>1\right]\notag\\
& =
\sum_{i=1}^{n+1}\E_{L} \left[\left. 1[X_i\in I]\cdot \left(\frac{1}{|I|}\sum_{j\in I} q_j - \frac{1}{|I|-1}\sum_{j\in I,j\neq X_i} q_j \right)\right| |I|>1\right] \tag{if $X_i\notin I$, $f(L)=f(L^{-i})$}\\
& =
\E_{L} \left[\left. \sum_{i=1}^{n+1}1[X_i\in I]\cdot \left(\frac{1}{|I|}\sum_{j\in I} q_j - \frac{1}{|I|-1}\sum_{j\in I,j\neq X_i} q_j \right)\right| |I|>1\right] \notag\\
& =
\E_{L} \left[\left. \sum_{k\in I}\left(\sum_{i=1}^{n+1} 1[X_i=k]\right)\cdot \left(\frac{1}{|I|}\sum_{j\in I} q_j - \frac{1}{|I|-1}\sum_{j\in I,j\neq k} q_j \right)\right| |I|>1\right] \notag
\end{align}
Since all elements of $I$ have the same number of successes, the sum $\sum_{i=1}^{n+1} 1[X_i=k]$ is the same for all $k\in I$ and since
\[
\frac{1}{|I|}\sum_{k\in I}\sum_{j\in I}q_j = \sum_{j\in I} q_j  = \frac{1}{|I|-1}\sum_{k\in I}\sum_{j\in I,j\neq k} q_j 
\]
this shows that 
\begin{equation}\label{eq:firstpart}
\sum_{i=1}^{n+1}\E_{L} \left[\left.f(L)-f(L^{-i}) \right| |I|>1\right]=0\,.
\end{equation}

Now let us consider the case $|I|=1$. 
Let $J\supseteq I$ denote the set of almost minimizers (i.e., which are either optimal or one away from being optimal).
We further condition on $J$ being some arbitrary fixed set $J_0$:
\begin{align}
&\E_{L} \left[\left.f(L)-f(L^{-(n+1)}) \right| |I|=1, J=J_0\right]\notag\\
&=\Pr\bigl[I= \{X_{n+1}\} \big\vert \lvert I\rvert = 1, J=J_0\}\bigr]\cdot \E_{L} \left[\left.f(L)-f(L^{-(n+1)}) \right| I=\{X_{n+1}\}, J=J_0\right]\tag{if $X_{n+1}\notin I$ then $f(L)-f(L^{-(n+1)})=0$}.
\end{align}
{Therefore, since $\Pr[I= \{X_{n+1}\}\vert \lvert I\rvert = 1, J=J_0\}]>0$, it is enough to consider
\begin{align}
&\E_{L} \left[\left.f(L)-f(L^{-(n+1)}) \right| I=\{X_{n+1}\}, J=J_0\right]\notag\\
&=\sum_{i\in J_0} \Pr\bigl(X_{n+1}=i \vert J=J_0, I=\{X_{n+1}\}\bigr)\cdot \E_{L} \left[\left.f(L)-f(L^{-(n+1)}) \right|X_{n+1}= i, I=\{X_{n+1}\}, J=J_0\right]\notag\\
&=\sum_{i\in J_0} \Pr\bigl(X_{n+1}=i \vert J=J_0, I=\{X_{n+1}\}\bigr)\cdot \E_{L}
\left[\left. q_i - \frac{1}{|J|}\sum_{j\in J}q_j \right|X_{n+1}= i, I=\{X_{n+1}\}, J=J_0\right]\notag\\
&=\sum_{i\in J_0} \Pr\bigl(X_{n+1}=i \vert J=J_0, I=\{X_{n+1}\}\bigr)\cdot \E_{L}
\left[\left. q_i - \frac{1}{|J_0|}\sum_{j\in J_0}q_j \right|X_{n+1}= i, I=\{X_{n+1}\}, J=J_0\right]\notag\\
&= \sum_{i\in J_0} \Pr\bigl(X_{n+1}=i \vert J=J_0,I=\{X_{n+1}\}\bigr)\cdot \bigl( q_i - \frac{1}{|J_0|}\sum_{j\in J_0}q_j\bigr) \notag\\
%
&= \sum_{i\in J_0} \frac{p_i}{\sum_{j\in J_0} p_j}\cdot \bigl( q_i - \frac{1}{|J_0|}\sum_{j\in J_0}q_j\bigr)\label{eq:13}\\
&= \frac{1}{\sum_{j\in J_0} p_j} \cdot 
\Bigl(\sum_{i\in J_0} p_iq_i - \frac{1}{|J_0|}\sum_{i,j\in J_0}p_iq_j \Bigr). \notag
\end{align}
Equation~\eqref{eq:13} above follows because the event that
{\it the set of almost minimizers $J$ is $J_0$ and the unique minimizer is $X_{n+1}$} is equivalent to the event that
{\it the set of minimizers with respect to the first~$n$ samples $X_1,\ldots, X_n$ is $J_0$ and $X_{n+1}\in J_0$.}}

Finally, to see that the above is non-positive we use Chebyshev's sum inequality, which asserts that if $a_1\leq\ldots  \leq a_n$ and $b_1\geq\ldots \geq b_n$ then $\frac{1}{n}\sum a_i\cdot \frac{1}{n}\sum b_i \geq \frac{1}{n}\sum a_ib_i$~\citep*{hardy1988inequalities}.
Thus, since $p_i=1-q_i$, this inequality implies that the left term inside the brackets is upper bounded by the right term, and so we get that the difference is non-positive. 
Hence, for every choice of~$J_0$:
\[
\E_{L} \left[\left.f(L)-f(L^{-(n+1)}) \right| |I|=1, J=J_0\right] \le 0\,,
\]
which together with Equation~\eqref{eq:firstpart} allows to conclude the proof of the lemma.
\end{proof}
This concludes the proof of Proposition~\ref{prop:multiclass}.
\end{proof}

\section{Wrapping Up}\label{sec:wrap}
\begin{proof}[Proof of Theorem~\ref{t:main}.]
Let $A$ be any learning algorithm
Proposition~\ref{prop:multiclass}, Proposition~\ref{prop:base-class}, 
    and Proposition~\ref{c:final} imply the existence of a monotone algorithm 
    $M$ such that for all $m \geq 2\cdot b(1) + b(0)$,
\begin{equation}\label{eq:final}
\E_{S\sim D^m}\Err(M(S)) \le \E_{S\sim D^m}\Err(A(S_{T-2})) + \cost(b(T-1)),    
\end{equation}
where
\begin{enumerate}
    \item $b(x)=4^x$, 
    \item $T=T(m)$ is the maximal integer such that $b(T-1) + \sum_{t=0}^{T-1} b(t)\le m$, and
    \item $S_{T-2}$ is an i.i.d sample from the source distribution $D$ of size $\sum_{t=0}^{T-2}b(t)$.
    \item $\cost(x) = 2\cdot\frac{1}{2\sqrt{x}} + 3\Bigl(\sqrt{{\ln(64 x)}/{x}} + \new{6\cdot\frac{36}{\sqrt{x}}}\Bigr)
        = O\Bigl(\sqrt{\frac{\log x}{x}}\Bigr)$,
\end{enumerate}

Since $M$ is monotone, it remains to prove that $M$'s performance is competitive with that of $A$.
The case of input sample-size $m< 2b(1) + b(0)=9$ is trivial. 
So, assume $m\geq 9$ and hence Equation~\ref{eq:final} holds.
By Items 1--4 above it suffices to show that $\lvert S_{T-2}\rvert \geq (m/30)-1$ and that
$b(T-1)\geq m/10$.
We proceed by showing that $T=T(m)=\Theta(\log m)$.
Recall that $T$ is the maximal positive integer such that
\[m \geq  4^{T-1} + \sum_{t=0}^{T-1}4^t = 2\cdot 4^{T-1} + \frac{4^{T-1}-1}{4-1} = \frac{7}{3}\cdot 4^{T-1}-\frac{1}{3}.\]
Thus,
\[T =1 + \Bigl\lfloor \log_4 \Bigl( \frac{3m+1}{7}\Bigr)\Bigr\rfloor, \]
and $b(T-1)$, $\lvert S_{T-2}\rvert$ satisfy:
\[b(T-1) = 4^{T-1} = 4^{\lfloor \log_4 ( \frac{3m+1}{7})\rfloor} \geq \frac{1}{4}\cdot\frac{3m+1}{7}\geq \frac{m}{10}.\]
\begin{align*}
\lvert S_{T-2}\rvert &= \sum_{t=0}^{T-2}4^t\\
                    &= \frac{4^{T-1}-1}{3}\\
                    &\geq \frac{\frac{1}{4}\cdot\frac{3m+1}{7} - 1}{3}\\
                    &=\frac{m}{28} - \frac{1}{14} \geq \frac{m}{30} - 1.
\end{align*}

\end{proof}

\section{Open Questions and Future Research}
We conclude this manuscript with some suggestions of open problems for future research.

\paragraph{Other Loss Functions.}
While the abstract framework developed in Section~\ref{sec:general} extends to other (bounded) loss functions, 
    the construction of the base classes $B_h$ is tailored to the zero/one loss.
    It will be interesting to explore to which loss functions can one extend Theorem~\ref{t:main}.

\paragraph{Can Monotone Learning Rules Achieve Optimal Rates?}
It will be interesting to explore whether the bound on the rate in Theorem~\ref{t:main} 
    can be strengthened to retain optimal learning rates.

For example, for PAC learnable classes $\mathcal{H}\subseteq\{0,1\}^X$, the optimal learning rate in the agnostic setting scales like $\sqrt{d/m}$, and in the realizable setting like $d/m$, where $d$ is the VC dimension and $m$ is the input-sample size.
    Can these optimal rates be achieved by a monotone learning rule? Note that Theorem~\ref{t:main} is off by a $\log m$ factor.

Another interesting setting to explore this question 
    is the model of \emph{universal} learning,
    in which one focuses on distribution-dependent rates~\citep*{Bousquet21Universal}.
    In contrast with the distribution-free nature of PAC learning,
    some classes can be learned exponentially fast,
    at a rate which scales like $\exp(-n)$~\citep*{schuurmans:97,Bousquet21Universal}.
    Can such classes be learned monotonically in this fast rate?

\paragraph{Monotone Empirical Risk Minimization.}
Which classes admit a monotone empirical risk minimizer (ERM)? 
One of the key technical steps in our proof was to show that the class $B_h$
admits a monotone ERM. Our proof exploited the symmetry of $B_h$ 
(specifically, that each example is classified correctly by exactly one hypothesis in $B_h$). It will be interesting to determine which other classes admit monotone ERMs.
In fact, as far as we know it is even open whether \emph{every} (learnable) class admits a monotone ERM.

\section{Acknowledgements}
We thank G\'{a}bor Lugosi for an insightful correspondence that helped to materialize the ideas leading to this work.
\new{We also thank Grigoris Velegkas and Amin Karbasi, and Gyeongwon Jeong and Jaehui Hwang for helping us correcting derivations in the proofs of Lemma~\ref{le:monmulti} and Lemma~\ref{le:prob}.}

\bibliographystyle{plainnat}
\bibliography{main.bib}

\begin{thebibliography}{16}
\providecommand{\natexlab}[1]{#1}
\providecommand{\url}[1]{\texttt{#1}}
\expandafter\ifx\csname urlstyle\endcsname\relax
  \providecommand{\doi}[1]{doi: #1}\else
  \providecommand{\doi}{doi: \begingroup \urlstyle{rm}\Url}\fi

\bibitem[Bousquet et~al.(2021)Bousquet, Hanneke, Moran, van Handel, and
  Yehudayoff]{Bousquet21Universal}
Olivier Bousquet, Steve Hanneke, Shay Moran, Ramon van Handel, and Amir
  Yehudayoff.
\newblock A theory of universal learning.
\newblock In Samir Khuller and Virginia~Vassilevska Williams, editors,
  \emph{{STOC} '21: 53rd Annual {ACM} {SIGACT} Symposium on Theory of
  Computing, Virtual Event, Italy, June 21-25, 2021}, pages 532--541. {ACM},
  2021.
\newblock \doi{10.1145/3406325.3451087}.
\newblock URL \url{https://doi.org/10.1145/3406325.3451087}.

\bibitem[Devroye et~al.(1996)Devroye, Gy{\"o}rfi, and
  Lugosi]{devroye1996probabilistic}
Luc Devroye, L{\'a}szl{\'o} Gy{\"o}rfi, and G{\'a}bor Lugosi.
\newblock \emph{A probabilistic theory of pattern recognition}, volume~31.
\newblock Springer Science \& Business Media, 1996.

\bibitem[Duda et~al.(2001)Duda, Hart, and Stork]{DudaHartStork01}
Richard~O. Duda, Peter~E. Hart, and David~G. Stork.
\newblock \emph{Pattern Classification}.
\newblock Wiley, 2 edition, 2001.
\newblock ISBN 978-0-471-05669-0.

\bibitem[Gu et~al.(2001)Gu, Hu, and Liu]{Gu01perf}
Baohua Gu, Feifang Hu, and Huan Liu.
\newblock Modelling classification performance for large data sets.
\newblock WAIM '01, page 317–328, Berlin, Heidelberg, 2001. Springer-Verlag.
\newblock ISBN 3540422986.

\bibitem[Hardy et~al.(1988)Hardy, Littlewood, and
  Pólya]{hardy1988inequalities}
G.~H. Hardy, John~E. Littlewood, and George Pólya.
\newblock \emph{Inequalities}.
\newblock Cambridge University Press, Cambridge, 1988.
\newblock ISBN 0521358809.

\bibitem[Lugosi(2002)]{lugosi2002pattern}
G{\'a}bor Lugosi.
\newblock Pattern classification and learning theory.
\newblock In \emph{Principles of nonparametric learning}, pages 1--56.
  Springer, 2002.

\bibitem[McDiarmid(1989)]{Mcdiarmid}
Colin McDiarmid.
\newblock On the method of bounded differences.
\newblock In \emph{Surveys in combinatorics, 1989 ({N}orwich, 1989)}, volume
  141 of \emph{London Math. Soc. Lecture Note Ser.}, pages 148--188. Cambridge
  Univ. Press, Cambridge, 1989.

\bibitem[Mhammedi(2021)]{mhammedi2021risk}
Zakaria Mhammedi.
\newblock Risk monotonicity in statistical learning.
\newblock \emph{Advances in Neural Information Processing Systems}, 34, 2021.

\bibitem[Pestov(2021)]{pestov2021universally}
Vladimir Pestov.
\newblock A universally consistent learning rule with a universally monotone
  error.
\newblock \emph{arXiv preprint arXiv:2108.09733}, 2021.

\bibitem[Schuurmans(1997)]{schuurmans:97}
D.~Schuurmans.
\newblock Characterizing rational versus exponential learning curves.
\newblock \emph{Journal of Computer and System Sciences}, 55\penalty0
  (1):\penalty0 140--160, 1997.

\bibitem[Shalev{-}Shwartz and Ben{-}David(2014)]{SSbook}
Shai Shalev{-}Shwartz and Shai Ben{-}David.
\newblock \emph{Understanding Machine Learning - From Theory to Algorithms}.
\newblock Cambridge University Press, 2014.
\newblock ISBN 978-1-10-705713-5.
\newblock URL
  \url{http://www.cambridge.org/de/academic/subjects/computer-science/pattern-recognition-and-machine-learning/understanding-machine-learning-theory-algorithms}.

\bibitem[Tax and Duin(2008)]{Tax08curves}
David Tax and Robert Duin.
\newblock Learning curves for the analysis of multiple instance classifiers.
\newblock volume 5342, pages 724--733, 12 2008.
\newblock ISBN 978-3-540-89688-3.
\newblock \doi{10.1007/978-3-540-89689-0_76}.

\bibitem[Viering et~al.(2019)Viering, Mey, and Loog]{viering2019open}
Tom Viering, Alexander Mey, and Marco Loog.
\newblock Open problem: Monotonicity of learning.
\newblock In \emph{Conference on Learning Theory}, pages 3198--3201. PMLR,
  2019.

\bibitem[Viering and Loog(2021)]{Viering21curves}
Tom~J. Viering and Marco Loog.
\newblock The shape of learning curves: a review.
\newblock \emph{CoRR}, abs/2103.10948, 2021.
\newblock URL \url{https://arxiv.org/abs/2103.10948}.

\bibitem[Viering et~al.(2020)Viering, Mey, and Loog]{viering2020making}
Tom~Julian Viering, Alexander Mey, and Marco Loog.
\newblock Making learners (more) monotone.
\newblock In \emph{International Symposium on Intelligent Data Analysis}, pages
  535--547. Springer, 2020.

\bibitem[Weiss and Battistin(2014)]{Weiss2014GeneratingWL}
Gary~M. Weiss and Alexander Battistin.
\newblock Generating well-behaved learning curves : An empirical study.
\newblock 2014.

\end{thebibliography}

\end{document}